\UseRawInputEncoding
\documentclass[10pt,twocolumn,letterpaper]{article} 

\usepackage{avss}
\usepackage{times}
\usepackage{epsfig}
\usepackage{graphicx}
\usepackage{amsmath}
\usepackage{amssymb}
\usepackage[scientific-notation=true]{siunitx}
\usepackage[pagebackref=true,breaklinks=true,letterpaper=true,colorlinks,bookmarks=false]{hyperref}

\avssfinalcopy % *** Uncomment this line for the final submission

\def\avssPaperID{97} % *** Enter the AVSS Paper ID here

\ifavssfinal\fi

\makeatletter
\def\ps@IEEEtitlepagestyle{%
  \def\@oddfoot{\mycopyrightnotice}%
  \def\@evenfoot{}%
}
\def\mycopyrightnotice{%
  {\footnotesize 978-1-6654-3396-9/21/\$31.00 \copyright 2021 IEEE\hfill}
  \gdef\mycopyrightnotice{}% just in case
}
\makeatother

\begin{document}

%%%%%%%%% TITLE
\title{Track Boosting and Synthetic Data Aided Drone Detection}

\author{Fatih Cagatay Akyon, Ogulcan Eryuksel, Kamil Anil Ozfuttu, Sinan Onur Altinuc \\
OBSS AI\\
OBSS Technology, Ankara, Turkey \\
{\tt\small \{fatih.akyon, ogulcan.eryuksel, anil.ozfuttu, sinan.altinuc\}@obss.com.tr}
%\and
%{\tt\small secondauthor@i1.org}
}

\maketitle
\thispagestyle{empty}

%%%%%%%%% ABSTRACT
\begin{abstract}
    This is the paper for the first place winning solution of the Drone vs. Bird Challenge, organized by AVSS 2021. As the usage of drones increases with lowered costs and improved drone technology, drone detection emerges as a vital object detection task. However, detecting distant drones under unfavorable conditions, namely weak contrast, long-range, low visibility, requires effective algorithms. Our method approaches the drone detection problem by fine-tuning a YOLOv5 model with real and synthetically generated data using a Kalman-based object tracker to boost detection confidence. Our results indicate that augmenting the real data with an optimal subset of synthetic data can increase the performance. Moreover, temporal information gathered by object tracking methods can increase performance further.
\end{abstract}

\let\thefootnote\relax\footnotetext{\mycopyrightnotice}

\section{Introduction}

Initially being used for military applications, the use of drones has been extended to multiple application fields, including traffic and weather monitoring \cite{elloumi2018monitoring}, smart agriculture monitoring \cite{tokekar2016sensor}, and many more \cite{shakhatreh2019unmanned}. Furthermore, with the COVID-19 pandemic, there has been a radical increase in the use of drones not only for autonomous delivery of essential grocery and medical supplies but also to enforce social distancing. Nowadays, small quadcopters can be easily purchased on the Internet at low prices, which brings unprecedented opportunities but also poses several threats in terms of safety, privacy, and security \cite{humphreys2015statement}.

The Drone vs. Bird Detection Challenge was launched in 2017, during the first edition of the International Workshop on Small-Drone Surveillance, Detection and Counteraction Techniques (WOSDETC) \cite{coluccia2019drone} as part of the 14th edition of the IEEE International Conference on Advanced Video and Signal based Surveillance (AVSS). This challenge aims to address the technical issues of discriminating between drones and birds \cite{coluccia2019drone}. Given their characteristics, in fact, drones can be easily confused with birds, particularly at long distances, which makes the surveillance task even more challenging. The use of video analytics can solve the issue, but effective algorithms are needed that can operate under unfavorable conditions, namely weak contrast, long-range, low visibility, etc.

To overcome these issues, firstly, we use synthetic data selectively to enrich the dataset. Secondly, we make use of the Kalman filter \cite{welch1995introduction} based object tracking method to track objects across time to eliminate false positives and enhance detection performance. Lastly, we propose a track boosting method for boosting the confidence scores of detections based on track statistics.

%------------------------------------------------------------------------- 
\section{Related Work}

In recent years, the application of deep learning-based detection methods has led to excellent results for a wide range of applications, including drone detection. However, due to the absence of large amounts of drone detection datasets, a two-staged detection strategy has been proposed in \cite{sommer2017flying}. First, the authors examined the suitability of different flying object detection techniques, i.e., frame differencing and background subtraction techniques, locally adaptive change detection, and object proposal techniques \cite{muller2017robust}, to extract region candidates in video data from static and moving cameras. In the second stage, a small CNN classification network is applied to distinguish each candidate region into drone and clutter categories.

In \cite{coluccia2021drone}, Gagné and Mercier (referred to as Alexis team) proposed a drone detection approach based on YOLOv3 \cite{redmon2018yolov3} and taking a single RGB frame as input. By integrating an image tiling strategy, this approach is able to detect small drones in high-resolution images successfully. Alexis Team leveraged the public PyTorch implementation of YOLOv3 with Spatial Pyramid Pooling (YOLOv3-SPP) made available by Ultralytics \cite{glenn_jocher_2020_3785397}. Spatial Pyramid Pooling \cite{he2015spatial} is a simple technique for which the input features are processed by pooling layers of different sizes in parallel and then concatenated to generate fixed-length feature vectors. Moreover, EagleDrone Team proposed a YOLOv5 based drone detection modality with a linear sampling-based data sub-sampling method. They propose to set the sampling probabilities using calculated loss per image. In addition to that, they utilize an ESRGAN based super-resolution technique to detect small and low-resolution drones.

\section{Proposed Technique}
\subsection{Detection Model}
The proposed technique focuses on a combination of two methods to improve the accuracy of drone detection performance using YOLOv5 \cite{yolov5} object detection method. YOLOv5 is selected because of its speed and performance on object detection tasks. In addition, it supports anchor optimization, which is proven to improve performance \cite{dvb2021} and feature pyramids \cite{lin2017feature} that handle objects at different scales.

\subsection{Synthetic Data}
\label{sec:synthetic}

\begin{figure}[!h]
\begin{center}
   \includegraphics[width=1\linewidth]{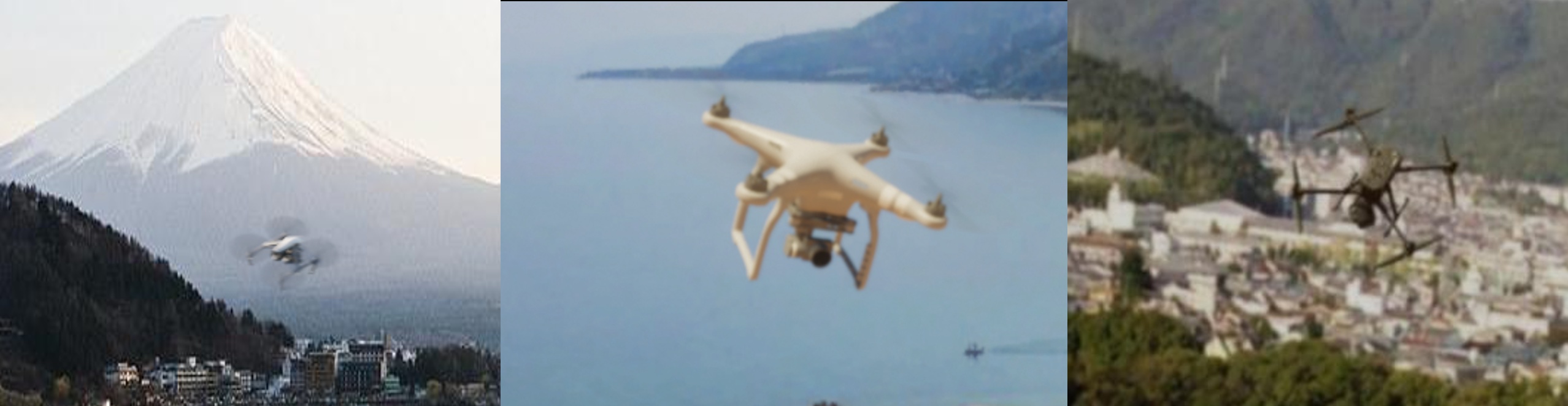}
\end{center}
   \caption{Samples from synthetically generated drones images.}
\label{fig:synthetic-drones}
\end{figure}

The use of synthetic data in deep learning appears helpful in scenarios where data is scarce or unavailable. Although synthetic data alone cannot show the same performance as real data, it has been seen that it increases performance when used alongside real data \cite{dvb2021}. Since there is no general method for creating synthetic data, each problem requires a unique approach. For the drone tracking problem, a method for creating labeled, randomized compositions by positioning 3D drone objects in front of 2D backgrounds were designed. This method was chosen because it is challenging to create a 3D randomized environment for the drone detection problem, and a location-independent object such as a drone can be used appropriately with 2D backgrounds. To generate the dataset, 3D drone models were rendered with various conditions such as position, rotation and lighting, and post-process effects on the randomized background images. Some samples from the synthetically generated dataset can be seen in Fig. \ref{fig:synthetic-drones}.

The most complex challenge encountered in the method is developing a solution to bridge the so-called "domain gap". To achieve this, experiments were conducted on the properties of the generated synthetic data, and the properties that make it similar to real data were investigated. In line with these studies, we created datasets with varying sample sizes that were optimized based on the features discovered. Finally, a series of experiments were conducted; first, a dataset was used with mixing synthetic and real data; second, a model trained on a synthetic dataset was used as a backbone for real data training.

\begin{table*}[!t]
    \centering
    \begin{tabular}{c|cccccc}
    Data & Technique & mAP & mAP-s & mAP-m & mAP-l \\
    \hline
    Combined Dataset & YOLOv5 + Tracker + Track Boosting & \textbf{79.4} & \textbf{86.2} & \textbf{72.7} & \textbf{70.3} \\
    Real Dataset & YOLOv5 + Tracker + Track Boosting & 76.1 & 86.6 & 67.6 & 43.0 \\
    Synthetic Dataset & YOLOv5 + Tracker + Track Boosting & 60.9 & 69.3 & 58.3 & 41.3 \\
    \hline
    Combined Dataset & YOLOv5 + Tracker & 78.8 & 85.7 & 71.6 & 65.3 \\
    Real Dataset & YOLOv5 + Tracker & 74.6 & 86.2 & 66.9 & 33.0 \\
    Synthetic Dataset & YOLOv5 + Tracker & 56.6 & 65.4 & 55.7 & 36.0 \\
    \hline
    Combined & YOLOv5 & 78.1 & 84.8 & 72.0 & 65.2 \\
    Real Dataset & YOLOv5 & 74.6 & 86.0 & 67.1 & 33.4 \\
    Synthetic Dataset & YOLOv5 & 55.6 & 63.0 & 55.6 & 35.2 \\
    \end{tabular}
    %\vspace*{-3px}
    \caption{Fine-tuning results for synthetic data augmentation and track boosting technique. In `Data` column, `Combined` means synthetic samples are used in training together with real data. In `Technique` column, `Y` means YOLOv5m6 model is used as detector and `T` means Kalman based tracker is applied on top of model detections. `mAP` corresponds to mean average precision at 0.50 threshold. mAP-s, mAP-m and mAP-l corresponds to small, medium and large object detection accuracies, respectively.}
    \label{tab:eval}
    %\vspace*{-15px}
\end{table*}

\subsection{Object Tracking And Tracker Based Confidence Boosting}

Object tracking algorithms are used to provide continuity of object detections over time.  While tracking the objects is not directly required, it provides temporal information about the objects in the video that can further improve performance.

\subsubsection{Tracker Method}
A simple Kalman-based tracking method is applied \cite{norfair} over the predictions of the object detection network. Kalman-based tracking uses a position and velocity-based process, object detection methods as measurements, and a hit counter mechanism. The tracking parameters are optimized for drone tracking with possibly moving cameras. 

\begin{figure}[!h]
\begin{center}
   \includegraphics[width=0.9\linewidth]{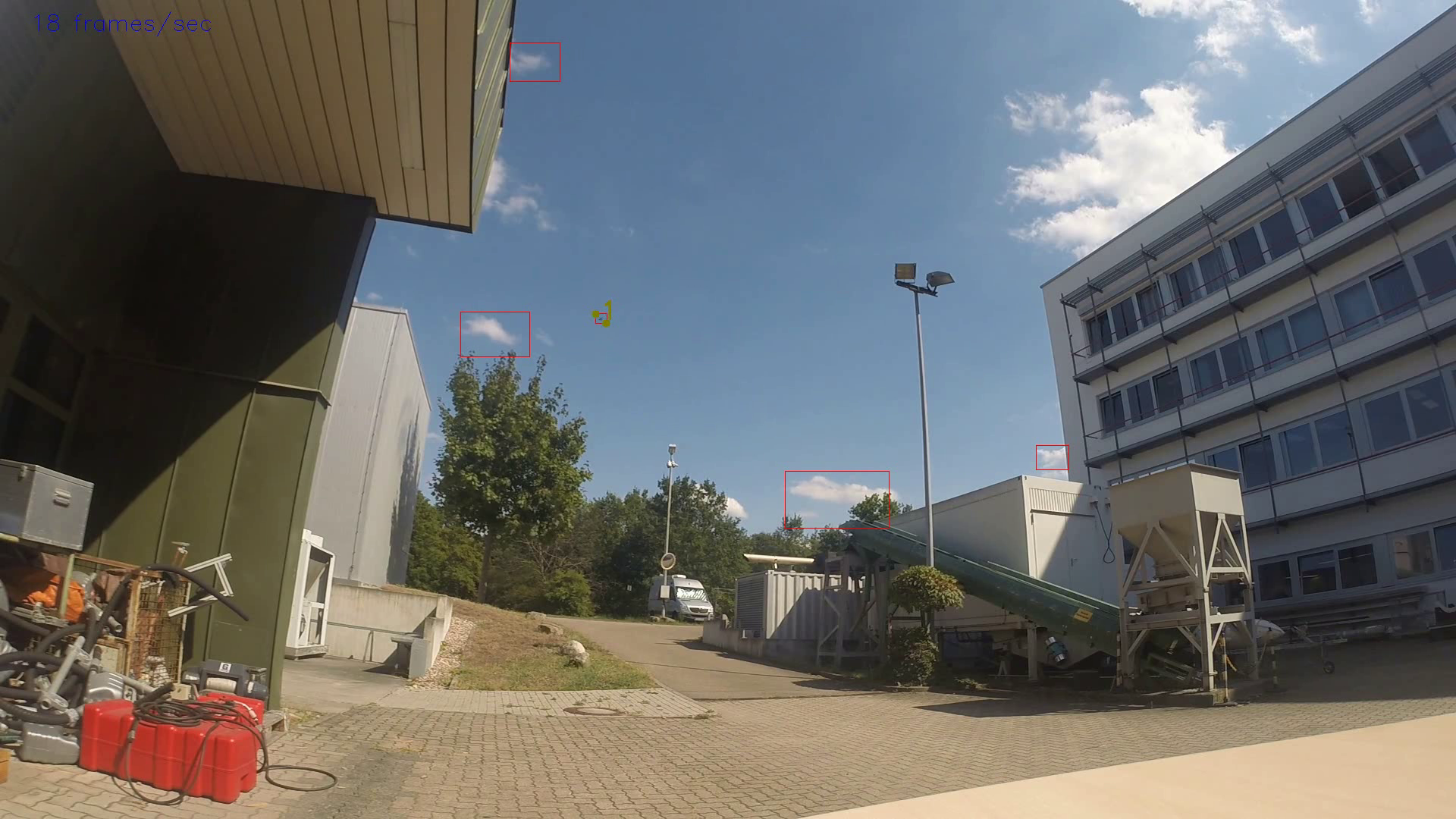}
\end{center}
   \caption{Tracker results and false positive detections on a frame. Drone in the frame has a tracking id of 1 that shows tracker tracks the frame. The other red boxes on the clouds are false positives that can be filtered out by the tracker.}
\label{fig:false-positives}
\end{figure}

One benefit of using a tracker system is that for a track to be formed successfully, the object detection model needs to provide a consistent stream of predictions close to the object's predicted location. Therefore false positives occurring at random positions usually fail to build up the necessary hit count to form a track, as shown in Fig. \ref{fig:false-positives}. This has a positive impact on the mAP score. 

\subsubsection{Track Boosting Method}
The tracks provide a spatiotemporal dimension and continuity over the predictions. This information is used in various ways to improve performance.

However, tracking can provide object predictions where the object detection method provides no predictions; tracker-only predictions are not very useful compared to detection predictions due to their low IoU rate and test set not having annotations for occluded objects. Furthermore, in conducted experiments, including the tracker predictions had a negative impact on the mAP score. Therefore only detections from the object detection method are used in the Tracker Boosting Method.

Also, object detection model confidence may vary significantly with moving objects as the object moves or changes orientation over time. With the track information provided by the tracking algorithm, we increased the confidence of the predictions within a track by averaging the max confidence score in the track with the confidence provided by the object detection algorithm as shown in  Eq. (\ref{eq:trackingboost}) where  ${ S'_{i,j} }$ is the score for a prediction with ${i}$ as track number and ${j}$ as the position in the track and ${s_i}$ is the vector of scores for track ${i}$.

\begin{equation}
\label{eq:trackingboost}
S'_{i,j}= {\frac{S_{i,j} + \max(s_i)}{2}}
\end{equation}

\begin{figure}[!h]
\begin{center}
   \includegraphics[width=0.9\linewidth]{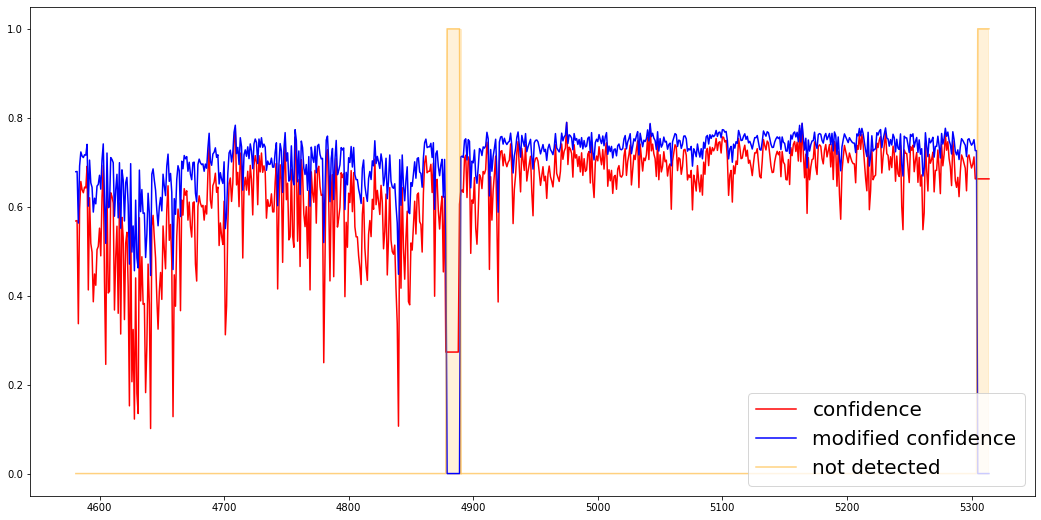}
\end{center}
   \caption{Plot of the confidence scores of a track. Red lines are the confidence scores of object detection and tracking. Blue lines are the results of the confidence increasing algorithm. In areas marked with orange, object detector failed to find the object.  }
\label{fig:conf-increase}
\end{figure}

\section{Experiments}
\subsection{Used Datasets}
\label{sec:datasets}

Before conducting the experiments, we randomly selected 15 of the videos from drone-vs-bird set \cite{dvb2021} as the validation set. Then extracted the frames from training and validation videos. We observed that using a subset of the training frames instead of the whole set prevents over-fitting resulting in improved accuracy. Therefore, uniformly sampled $10^4$ frames from the drone vs. bird training is used as training data in experiments. This dataset will be referred as "Real Dataset". 

Synthetic datasets with different features were generated using the method mentioned in sec. \ref{sec:synthetic}.These datasets with special features are:

\begin{description}
  \item[$\bullet$ Original] Original rendered image without spesific features
  \item[$\bullet$ Noise] Image rendered with film grain noise post processing effect
  \item[$\bullet$ Optimal drone sizes] Drones sized according to normal distribution  
  \item[$\bullet$ Blur] Rendered image with Gaussian Blur optimized for backgrounds
\end{description}

The dataset generated with $10^4$ images by optimizing these features is called "Synthetic Dataset".

To perform experiments in which synthetic data will be used alongside real data, we created a combined dataset of \num[round-precision=3, round-mode=figures]{10500} images which will be referred as "Combined Dataset". This dataset includes all $10^4$ samples from "Real Dataset" and an optimal sub-sample of 500 images from 'Synthetic Dataset'. The reason to include just a small subset of the synthetic images is to avoid the domain gap mentioned in sec. \ref{sec:synthetic}.

Moreover, we have also trained models with a combined dataset containing images from mav-vid \cite{rodriguez2020adaptive} and realworld-uav \cite{9205392} drone detection datasets. However, since their distributions differ from the drone-vs-bird dataset, our drone-vs-bird validation accuracy dropped, so we do not mention combined dataset training results in this paper.

\subsection{Results}

\begin{figure}[!h]
\begin{center}
   \includegraphics[width=1\linewidth]{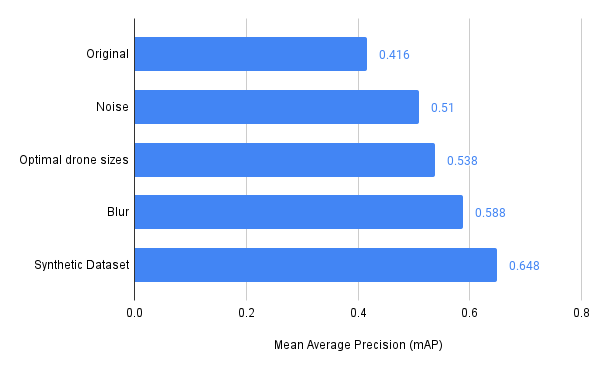}
\end{center}
   \caption{Evaluation result for different synthetic features}
\label{fig:syntheti_map_chart}
\end{figure}

Firstly, a series of experiments were performed on various synthetic datasets to find the optimal synthetic features. All experiments were performed with COCO pretrained YOLOv5m6 model of 1333 input image size, 8 batch size, and 10 epochs setup and with datasets described in \ref{sec:datasets}. For data sampling, inference and evaluation, our open source vision package SAHI \cite{akyon2022sahi} is utilized. As seen in Fig. \ref{fig:syntheti_map_chart}, experiments revealed that the most important property is the amount of blur applied to the image. In addition, it was understood that the distribution of drone sizes and noise effect similar to film grain significantly affect the performance. The results of the synthetic data experiments are used to create the optimized synthetic dataset known as 'Synthetic Dataset'. The training results obtained with the 'Synthetic Dataset' show that with the correct features, a training set consisting of only synthetic images gives us acceptable results considering no real images were required in the process.

For the real data (drone-vs-bird) experiments, COCO pretrained YOLOv5m6 model is fine-tuned on drone-vs-bird training split with 1333 input image size, 6 batch size for 10 epochs. During inference, vanilla YOLOv5 detection results are taken as a baseline. Then Kalman filter-based tracker is applied on top of model detections. Norfair package \cite{norfair} is utilized for the Kalman filter implementations with measurement (R) and process (Q) uncertainty parameters of 0.2 and 1, respectively. Lastly, the track boosting technique is applied to the tracker output for further performance improvement.

As seen in Table \ref{tab:eval}, real data gives better results than synthetic data; however, augmenting real data with synthetically generated data improves the validation results by up to 4.2 AP in all scenarios. Moreover, by applying a Kalman filter-based tracker, base results can be improved by up to 1 AP. More importantly, applying the track boosting method on top of a tracker provides us with an additional 1.5 and 0.6 AP improvement in real and combined dataset experiments, respectively.

\section{Conclusion}
Our results show that a YOLOv5 model fine-tuned only on synthetically generated images can achieve acceptable performance on drone detection tasks. Moreover, mixing an optimal subset of synthetic data yielded much better results than using real and synthetic images by themselves. Usage of the tracker improves upon the object detection performance in all cases. This improvement may be a result of filling out missing frames and eliminating the false positives by the tracker's internal mechanism. These results can be further improved by adjusting the frame predictions using the track information. Using the maximum confidence value in a track as a reference value, the overall mAP score increased. In cases where both the tracking algorithm and the object detection provide a prediction using the prediction from the object detection model results in improved accuracy as well.

{\small
\bibliographystyle{ieee}
\bibliography{egbib}
}

\end{document}